\documentclass[letterpaper, 10 pt, conference]{ieeeconf} 
\IEEEoverridecommandlockouts
\usepackage[font=small]{caption}
\let\labelindent\relax
\usepackage[]{amsmath,amssymb,dsfont,color,graphicx,subcaption,mathtools,nccmath,bm,enumitem}
\usepackage{url}

\usepackage[space]{cite}  

\renewcommand{\baselinestretch}{0.965}
\begin{document}

\title{\LARGE \bf Motion Planning for a Climbing Robot with Stochastic Grasps}

\author{Stephanie Newdick$^1$, Nitin Ongole$^1$, Tony G. Chen$^2$, \\ Edward Schmerling$^1$, Mark R. Cutkosky$^2$, Marco Pavone$^1$ 
\vspace{-10pt}
\thanks{$^1$Stephanie Newdick, Nitin Ongole, Edward Schmerling, and Marco Pavone are with the Department of Aeronautics and Astronautics, Stanford University, Stanford, CA 94305. \{{\tt snewdick, nitin99, schmrlng, pavone}\}{\tt@stanford.edu}.}
\thanks{$^2$Tony G. Chen and Mark R. Cutkosky are with the Department of Mechanical Engineering, Stanford University, Stanford, CA 94305. \{{\tt agchen, cutkosky}\}{\tt@stanford.edu}.}
}

\maketitle
\thispagestyle{empty}
\pagestyle{empty}

\begin{abstract}
Motion planning for a multi-limbed climbing robot must consider the robot's posture, joint torques, and how it uses contact forces to interact with its environment. This paper focuses on motion planning for a robot that uses nontraditional locomotion to explore unpredictable environments such as martian caves. Our robotic concept, ReachBot, uses extendable and retractable booms as limbs to achieve a large reachable workspace while climbing. Each extendable boom is capped by a microspine gripper designed for grasping rocky surfaces. 
ReachBot leverages its large workspace to navigate around obstacles, over crevasses, and through challenging terrain. 
Our planning approach must be versatile to accommodate variable terrain features and robust to mitigate risks from the stochastic nature of grasping with spines. In this paper, 
we introduce a graph traversal algorithm to select a discrete sequence of grasps based on available terrain features suitable for grasping.
This discrete plan is complemented by a decoupled motion planner that considers the alternating phases of body movement and end-effector movement,
using a combination of sampling-based planning and sequential convex programming to optimize individual phases. 
We use our motion planner to plan a trajectory across a simulated 2D cave environment with at least 95\% probability of success and demonstrate improved robustness over a baseline trajectory.
Finally, we verify our motion planning algorithm through experimentation on a 2D planar prototype.
\end{abstract} %

\section{Introduction} \label{sec:intro}
There is a growing interest in planetary operations that require robots capable of locomotion and mobile manipulation in a variety of terrains. For example, NASA's interest in exploring caves, cliffs, and other rocky terrain on Mars~\cite{NRC2011,LapotreORourkeEtAl2020} motivates the need for space robots that combine sparse-anchored mobility with high-wrench (force and moment) manipulation. However, there is a key technology gap in existing solutions: small robots are typically restricted to a small reach and limited wrench capability. Conversely, large robots, particularly rigid-link articulated-arm robots, are hampered by high mass and complexity, which scale poorly with increased reach.

\begin{figure}[t!]
    \centering
	\includegraphics[width=0.98\columnwidth]{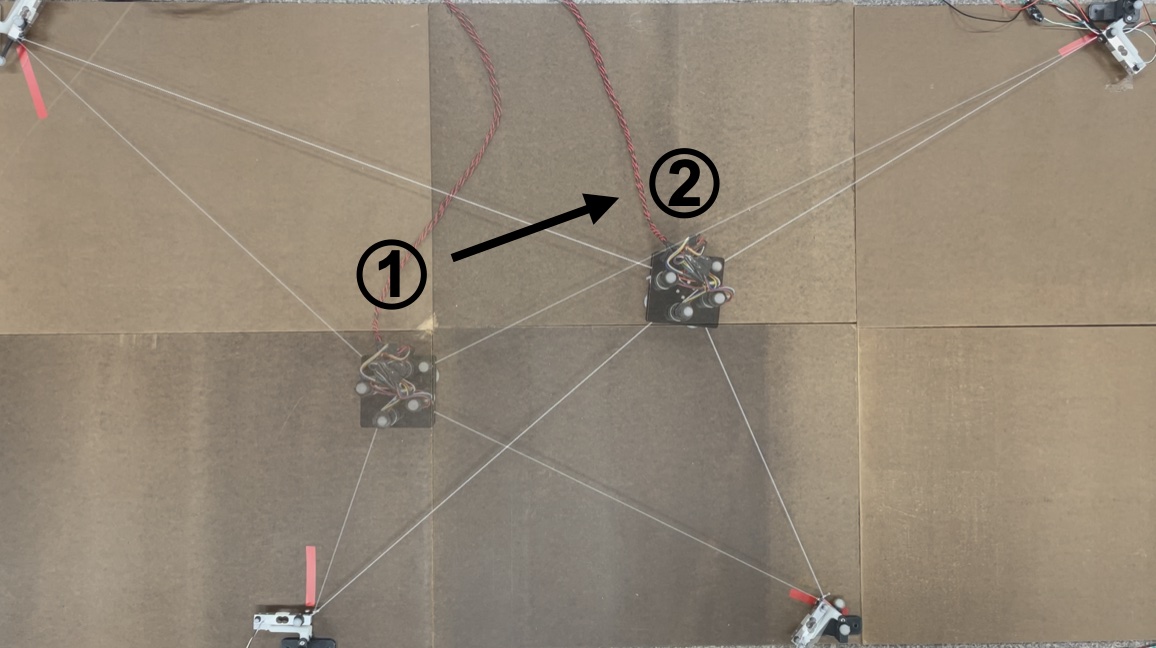}
    \caption{The experimental platform for ReachBot consists of a compact body with elastic cords in place of extendable booms. ReachBot uses force control to move its body while maintaining contact forces within the limit surfaces of all four anchor points.}
    \label{fig:hardware}
    \vspace{-15pt}
\end{figure}

ReachBot is a new robot concept for mobile manipulation in environments that require climbing, particularly with sparse anchor points.
As debuted in our past works~\cite{SchneiderBylardEtAl2022,ChenMillerEtAl2022}, ReachBot consists of a small body that uses multiple extendable booms as appendages to reach to nearby or distant rock surfaces and grip them.
These booms are ideal for space applications~\cite{LeclercWilsonEtAl2017}: they are lightweight and compact when rolled up, while still being strong when deployed and capable of extending many times the span of the robot body (Fig.~\ref{fig:hardware}).

With a large reachable workspace, ReachBot can target distant anchor points to pass over local irregularities, leveraging the advantages of a large robot (e.g.,~RoboSimian,
Tri-\textsc{ATHLETE}~\cite{HebertBajracharyaEtAl2015,Wilcox2012,HeverlyMatthewsEtAl2010}) but with reduced weight and complexity.
Climbing robots of similar weight (e.g.,~\textsc{LEMUR}~\cite{Parness2017}) have limited reach, preventing them from traversing environments with sparse anchor points or obstructions such as rockfall and crevasses.
ReachBot's large workspace allows for versatility in anchor point selection, a notable advantage when navigating these environments.
The dynamical model and simulation presented in past work demonstrated ReachBot's feasibility~\cite{SchneiderBylardEtAl2022}, but did not provide a high-level motion plan for ReachBot to explore and adapt to unique terrain geometries.
In this paper we present a planning approach that leverages ReachBot's versatile design while accounting for the difficult terrain encountered in a martian cave. 



ReachBot’s motion planner must address multiple challenges that arise from its mobility paradigm. It uses microspine grippers to
create attachments to surrounding basaltic surfaces, allowing it to manipulate its body by applying forces~\cite{ChenMillerEtAl2022}. The stochastic nature of grasping rock with spines introduces two complications. First, the non-zero failure probability of each new grasp highlights the importance of making as few regrasps as possible, requiring a motion planner that looks ahead at least a few moves before grasping a particular anchor point. Second, the stochasticity of the limit surface in force space for each attachment introduces uncertainty into the planning that we must consider to ensure robust mobility. Furthermore, the dynamic and kinematic constraints create a highly nonlinear system.

\textit{Statement of Contributions:}
This paper addresses the need for a grasp-aware motion planner to generate robust trajectories for an extendable-boom climbing robot through three main contributions. 
(1) We propose ReachBot Planner (RBP), a framework for non-gaited climbing robot that splits the motion planning problem into three decomposed segments: footstep planning, body movement, and end-effector movement.
(2) We introduce a convex metric for stochastic grasp success and use this metric to quantify and optimize the robustness of the planned trajectories. 
(3) We leverage the convexity of this metric and employ sequential convex programming to locally optimize trajectories between discrete footsteps, then validate the trajectories in simulation and on a 2D prototype.

\textit{Paper Organization:} 
The rest of this paper is organized as follows. In Section~\ref{sec:related}, we discuss existing motion planning approaches for legged and climbing robots. In Section~\ref{sec:problem_formulation}, we introduce our robotic formulation and gripper model and present metrics for robustness and feasibility of the system. In Section~\ref{sec:motion_planning}, we present a hierarchical motion plan wherein we first determine a discrete sequence of footsteps, then use sampling-based planning and sequential convex programming to optimize each continuous movement segment with respect to the metric of robustness. In Section~\ref{sec:results}, we compare our strategy against a naive baseline in simulation and validate our simulation results on a hardware prototype.

\section{Related Work} \label{sec:related}
Motion planning for legged or climbing robots requires a combination of discrete footstep planning and continuous trajectory optimization. 
Many recent approaches use mixed-integer convex optimization methods to find the footstep and continuous trajectory simultaneously~\cite{Aceituno-CabezasMastalliEtAl2017, AhnChaeEtAl2018}. Additionally, several papers have demonstrated the ability of nonlinear solvers to provide coupled motion plans~\cite{WinklerBellicosoEtAl2018}. However, such methods have proven to be prohibitively slow for robots with highly nonlinear dynamics like ReachBot~\cite{LinZhangEtAl2019}.

Alternatively, decoupled planners for multi-contact problems offer less optimal but faster solutions. Decoupled planners can be split into two categories: motion-before-contact and contact-before-motion. The former approach uses continuous planning to find a centroid trajectory, then determines contact placement. It is effective for robots using gaited mobility or for robots locomoting on flat or otherwise predictable terrain where contact points will always be available ~\cite{PosaCantuEtAl2014, LaurenziHoffmanEtAl2018, SintovAvramovichEtAl2011}. Conversely, a contact-before-motion approach is required for robots such as ReachBot that navigate uncertain terrain with irregular and unpredictable geometry. 

Contact-before-motion planners generate a discrete plan by considering adjacent contacts, often using graph search methods. After a sequence of contacts is determined, a continuous trajectory is planned to string the sequence together using optimal control or sampling-based planning~\cite{HauserBretlEtAl2005}.  Contact-before-motion planners typically use optimal or locally optimal control methods such as sequential convex programming (SCP) to minimize control effort~\cite{SchulmanDuanEtAl2014}.

Capuchin~\cite{ZhangLatomb2013} and LEMUR-3~\cite{Parness2017} offer two examples of non-linear systems designed for climbing in irregular terrain for which contact-before-motion planners have been demonstrated. The motion planner for Capuchin generates a footstep path by traversing a graph of stances where every feasible transition is an edge on the graph; LEMUR's sequence of stances is chosen by a human operator. Both robots use sampling-based methods to compute feasible paths between two stances.
Although many robots rely on contact interactions that are either unknown or inherently stochastic, existing motion planners do not explicitly account for this uncertainty. Due to the stochastic nature of ReachBot's grasps, RBP must account for uncertainty to remain robust to attachment failure. Our contribution is therefore to present a contact-before-motion planner for a multi-contact robot with nonlinear dynamics that minimizes the likelihood of failure.

\section{Problem Formulation} \label{sec:problem_formulation}
In this section, we reintroduce a formulation of ReachBot that will be used throughout this work.
We then introduce a grasp limit surface model that represents the stochasticity of grasping with microspines as Gaussian uncertainty. By parameterizing ReachBot's probability of success in any configuration, this grasp model is crucial to the development of a robust motion plan.
Finally, we define a notion of transition feasibility used to link sequential footsteps.

\begin{figure}[tp]
    \centering
    \includegraphics[width=1.0\linewidth]{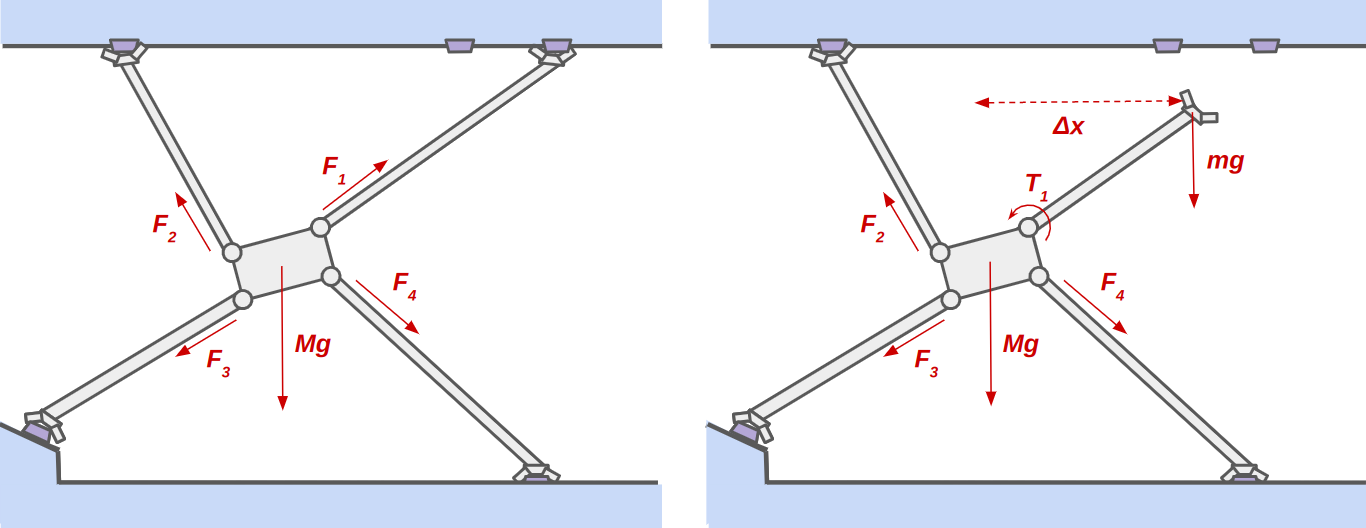}
    \caption{(Left) ReachBot in a 4-stance with all four booms attached. The axial forces $F_i$ for each boom $i$ must counteract the weight of the body, $Mg$. (Right) ReachBot in a 3-stance where one end-effector is detached. The weight of the gripper provides an external torque ($m g\Delta x $) on the body which must be supported by the active shoulder joint and the other three booms.}
    \label{fig:reachbot_model}
    \vspace{-10pt}
\end{figure}

This paper assumes a 2D adaptation of ReachBot to coincide with past work in modeling and prototype development~\cite{SchneiderBylardEtAl2022,ChenMillerEtAl2022}.
Planar ReachBot includes a robot body with four 
shoulder mechanisms, each having two controllable joints: (1) a prismatic joint composed of a motor that extends and retracts a deployable boom (see $F_i$ in Fig.~\ref{fig:reachbot_model}) and (2) a revolute shoulder joint that pivots the boom in the plane. As in our past work, ReachBot moves by alternating between body movement and end-effector movement. A \textit{stance} is an assignment of end-effectors to anchor points. During body movement, shown in Fig.~\ref{fig:reachbot_model} (left), all four of ReachBot’s end-effectors remain anchored in a ``4-stance" while the body
repositions. During end-effector movement, ReachBot must hold itself in a ``3-stance" while it moves its free end-effector to a new anchor point as shown in Fig.~\ref{fig:reachbot_model} (right). 

\subsection{Gripper model}
The contact model presented in this section determines the allowable pulling forces (magnitude and direction) for a given grasp.
Microspine grippers have been demonstrated to sustain large loads on a variety of rock-climbing robots ~\cite{Parness2017,AsbeckKimEtAl2006,RuotoloRoigEtAl2019}. Past analyses of spiny grippers developed models to generate a stochastic limit surface for a given gripper and surface topography~\cite{JiangWangEtAl2018,WangJiangEtAl2019}. We are currently developing models to construct a 6D wrench limit surface for ReachBot's gripper based on grasp angle and the size and distribution of asperities in the rock. For this work, we approximate this limit surface in force space as a stochastic half-ellipsoid parameterized by a semi-major and semi-minor axis.

\begin{figure}[tp]
    \centering
    \begin{subfigure}[]{0.26\textwidth}
	\includegraphics[width=\textwidth]{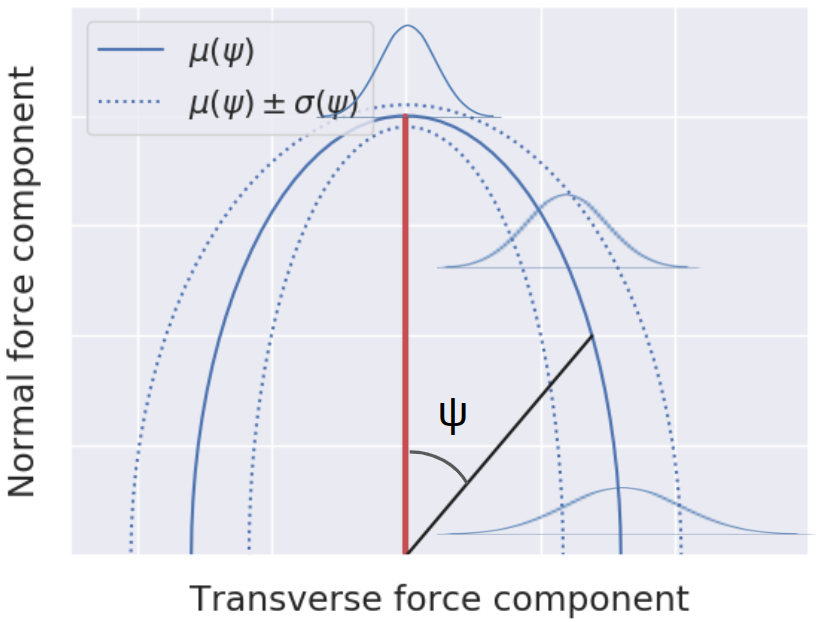}
	\label{subfig:limit_surface}
    \end{subfigure}
    \begin{subfigure}[]{0.2\textwidth}
	\includegraphics[width=\textwidth]{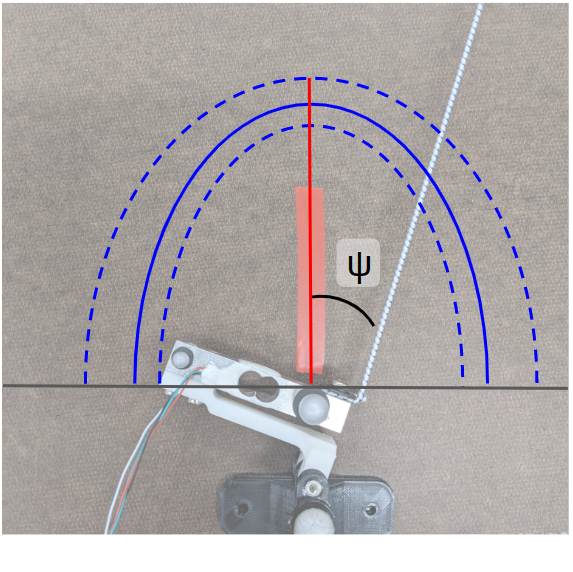}
	\label{subfig:limit_surface_hardware}
	\end{subfigure}

    \caption{(Left) Slice of stochastic half-ellipsoid used to model the limit surface of a grasp. Direction of pulling force is parameterized by $\psi$, where $\psi=0$ corresponds to a purely normal force, and $\psi=\pm\pi/2$ denotes lateral pulling. The expected limit is given by $\mu(\psi)$, and uncertainty is taken to be Gaussian, with standard deviation $\sigma(\psi)$ varying linearly with $\psi$. (Right) Limit surface ellipsoid overlaid on prototype anchor. We determine the limiting force as a function of $\psi$, then compare it to the pulling force measured by a strain gauge.}
    \label{fig:limit_surface}
        \vspace{-5pt}
\end{figure}

Fig.~\ref{fig:limit_surface} shows a 2D slice of an example ellipsoidal limit surface. The force that the grasp can sustain is a function of the pull angle, given by $\psi$. The spine-surface interaction is itself a probabilistic function~\cite{JiangWangEtAl2018}, and therefore the expected sustainable gripper force at a given angle $\psi$ is also stochastic. We assume that this uncertainty can be modeled as Gaussian, with the standard deviation increasing linearly as a function of $\psi$. 
This is consistent with Gaussian models of asperity slope variation on basaltic surfaces~\cite{Jiang2017}. Asperity slope will be the largest source of uncertainty for ReachBot's gripper design, so we assume an overall Gaussian uncertainty model.

\subsection{Robustness metrics}
ReachBot's robustness to failure depends on its configuration, defined by stance, pose, and contact forces.
In this section, we define a metric for configuration robustness based on the likelihood of failure of each individual grasp.

We consider a generalized ReachBot with $n$ booms, so the formulation can be applied to a 3-stance, 4-stance, or future $n$-stance 3D configuration. Since the wrist joints are passive, the pull angle $\psi_{\{i\}}$ for each grasp $i$ is fully determined by the anchor point coordinates and the pose and orientation of ReachBot.
Using a Gaussian model of uncertainty, the maximum sustainable force by grasp $i$ is given by
\begin{equation}
    f_{{\{i\}},\text{max}} \sim \begin{mathcal}{N}\end{mathcal} \big(\mu_{\{i\}}(\psi_{\{i\}}), \sigma_{\{i\}}^2(\psi_{\{i\}})\big),
\end{equation}
where $\begin{mathcal}{N}\end{mathcal}(\mu_{\{i\}}, \sigma_{\{i\}})$ is the stochastic limit surface for grasp $i$. We define the probability of success for the $n$-stance configuration as the joint probability that all forces $f_{\{i\}}$ are within their respective limit surfaces. 
Since the probability of each grasp's success is independent, we can apply conditional independence to write the joint probability as
\begin{equation}
    P \big( f_{\{1:n\}} \le f_{\{1:n\},\text{max}} | \psi_{\{1:n\}} \big) = 
    \prod_{i=1}^n P \big(f_{\{i\}} \le f_{\{i\},\text{max}} | \psi_{\{i\}} \big).
\end{equation}
By negating the force terms and applying the Gaussian cumulative distribution function $g(\cdot)$, we can rewrite the probability of success as
\begin{equation}
  \begin{aligned}
    P \big(-f_{\{1:n\},\text{max}} \le -f_{\{1:n\}} | \psi_{\{1:n\}} \big) = \\
    \prod_{i=1}^n g\bigg(\frac{-f_{\{i\}} + \mu_{\{i\}}(\psi_{\{i\}})}{\sigma_{\{i\}}(\psi_{\{i\}})}\bigg).
    \end{aligned}
\end{equation}
We take the $\log$ of both sides to define a convex robustness function
\begin{equation}
    \label{eq:r_prob}
    r(s,u) = 
    \sum_{i=1}^n \log \bigg[ g\bigg(\frac{-f_{\{i\}} + \mu_{\{i\}}(\psi_{\{i\}})}{\sigma_{\{i\}}(\psi_{\{i\}})}\bigg) \bigg],
\end{equation}
where $r(s,u)$ represents the negative $\log$ of the probability of success for a stance with a configuration parameterized by state vector $s$ and control vector $u$.

For the current robot prototype, the gripper will reach failure before enough pulling force can be applied to break the boom in tension. Therefore, the robustness of a configuration depends only on the probability of grasp failure. In future work, our gripper will also be able to apply modest pushing force on the wall, so robustness will additionally depend on potential failure due to exceeding the Coulomb friction cone, or the compression and bending limits of the boom.

Even with adequate force sensing capability and grasp limit surface modeling, the stochastic nature of spine contact on unknown rock surfaces introduces a non-zero chance of failure with every new grasp.
Moreover, there is some chance that the rock could fracture.
Therefore, to increase the robustness of a multi-footstep trajectory, we must additionally aim to minimize the overall number of regrasps. 

\subsection{Transition feasibility} \label{sec:feasibility}
To define transition feasibility, or the feasibility of moving between two stances, we must first define the feasibility of a pose. For a given stance, ReachBot's pose (position and orientation) is feasible if it meets the following requirements:
\begin{itemize}[leftmargin=15pt]
    \item It meets kinematic constraints. For our prototype, the booms can retract to a minimum of $1$cm and extend to a maximum of $2$m. The shoulder joints have a range of motion of $\pm 45^\circ$ in either direction from equilibrium. 
    \item No part of the robot body or booms is colliding with itself or with the environment.
    \item There exists a combination of control inputs where ReachBot resists external forces (e.g. gravity) \textit{and} the configuration's robustness~\eqref{eq:r_prob} is greater than some minimum value. For the rest of this paper, we define a minimum allowable probability of success of $95\%$, so
    \begin{equation}
        r_\text{min} = -\log(0.95).
    \end{equation}
\end{itemize}

A transition between two 4-stances is feasible if there exists a pose that is (1) feasible in the starting 4-stance, (2) feasible in the goal 4-stance, (3) feasible in the common 3-stance while holding the free boom statically at its starting anchor point, and (4) feasible in the common 3-stance while holding the free boom statically at its final anchor point. In other words, for a stance transition to be feasible, there must exist a pose that ReachBot can hold throughout the entire maneuver while maintaining robustness of at least $r_\text{min}$.



	

\section{Robust Motion Planning} \label{sec:motion_planning}
In this section, we present ReachBot Planner decomposed in three pieces: footstep planning, body movement, and end-effector movement. First, we present a discrete footstep planner that uses graph search to define a sequence of stances linked by feasible transitions. From this discrete plan, we break ReachBot's motion down into a sequence of alternating body movement and end-effector movement phases. 
For both of these continuous phases, we generate a feasible trajectory using sampling-based techniques, then locally optimize these trajectories using sequential convex programming.
We present representative samples of a full path generated by RBP for a simulated 2D environment where the planner has a priori knowledge of the randomly-generated anchor point locations and parameters of the limit surfaces' ellipsoids.

\begin{figure}[h]
    \centering
	\includegraphics[width=0.75\columnwidth]{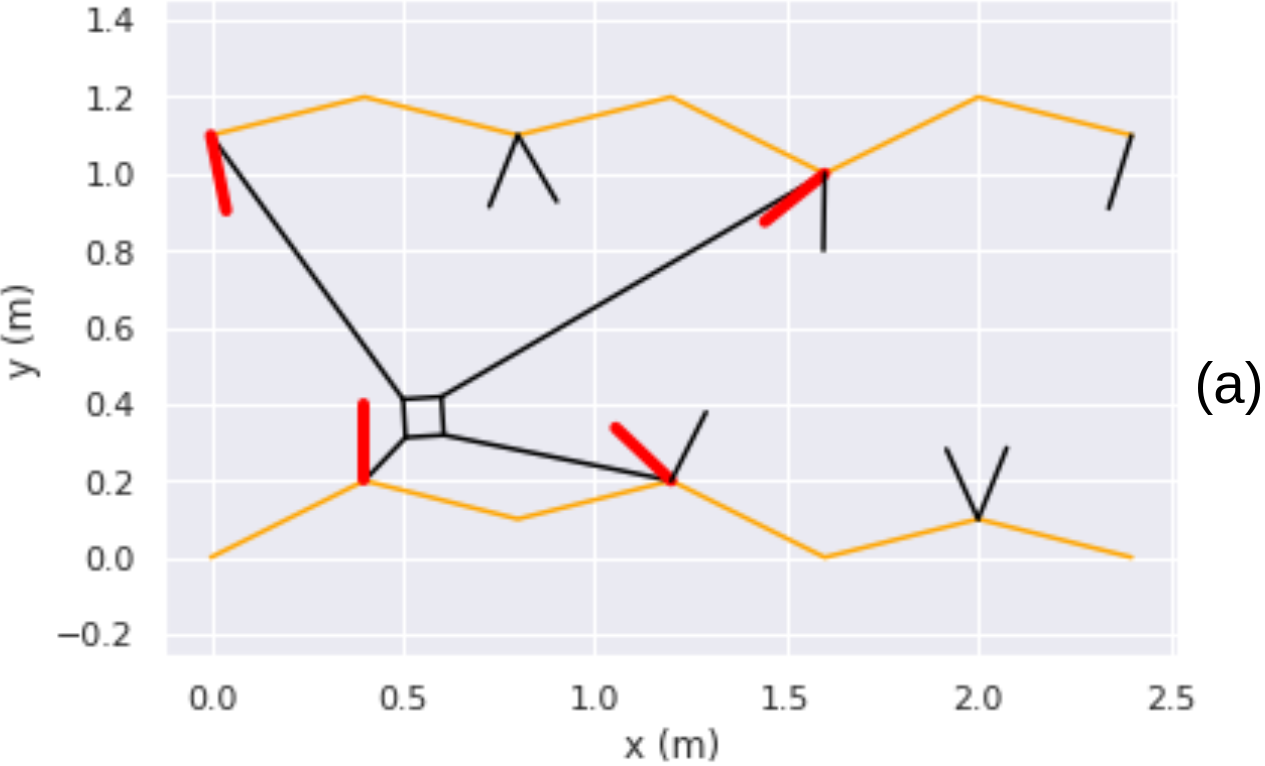}
	\includegraphics[width=0.75\columnwidth]{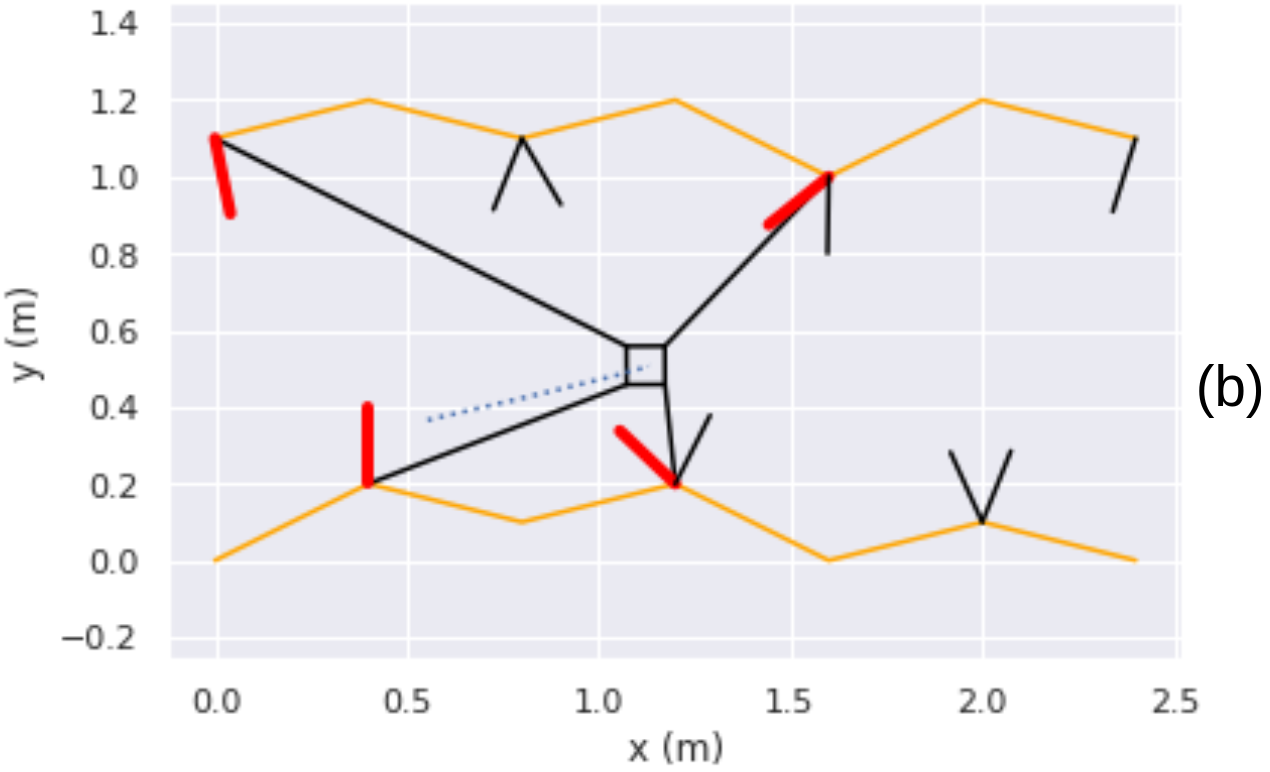}
	\includegraphics[width=0.75\columnwidth]{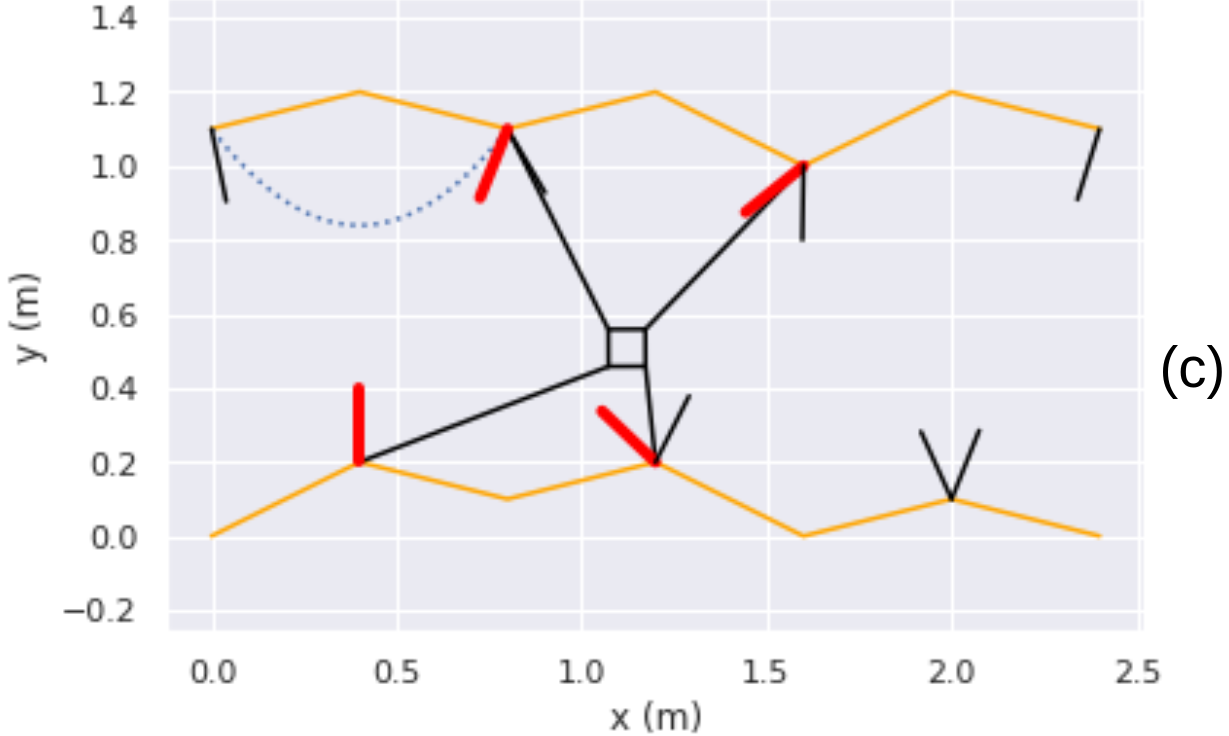}
	
    \caption{Snapshots of ReachBot's configuration at discrete points in its footstep plan. The solid orange lines represent the cave walls, and the black line segments represent the line at $\psi=0$ (see Fig.~\ref{fig:limit_surface}) for each respective anchor point limit surface. This line is red if that anchor point is being grasped. Figure (a) shows a snapshot of a 4-stance before body movement. From (a) to (b), ReachBot undergoes a body movement phase to assume a configuration where it can stably detach and move its top left end-effector. From (b) to (c), ReachBot moves its end-effector between adjacent 4-stances. The blue dotted line signifies the trajectory of either movement.}
    \label{fig:level2}
    \vspace{-0pt}
\end{figure}

\subsection{Footstep planning}\label{sec:level2}
The overall goal of footstep planning is to find a path of feasible poses and stance transitions to move the body from its starting position to its goal position. We assume for now that we have full knowledge of the available grasp points, including their position, orientation, and limit surface parameters throughout the environment.

We construct a graph wherein each 4-stance is a vertex, and each edge connecting adjacent vertices represents the common 3-stance. We use A* to traverse the graph with distance-to-goal heuristic to encourage larger steps and therefore fewer regrasps. To speed up graph traversal, we only populate edges where there is a feasible transition between the two vertices. The limiting torques that a 3-stance must resist during a transition occur immediately after detachment and immediately before initiating the next grasp. These two scenarios represent the maximum horizontal extension of the moving boom in either direction, and therefore the maximum external moments exerted on the body.
We use random sampling with local optimization to search for a feasible pose that can resist both limiting torques. If such a point is found, the transition is feasible, and we continue searching the graph in that direction. 
Fig.~\ref{fig:level2} shows an example of two steps ReachBot takes during its path from its start to goal state. 
%
\subsection{Body movement phase}\label{sec:level1}
In the body movement phase, ReachBot moves within a 4-stance to prepare for the next stance transition. The footstep path guarantees that each pose at the beginning and end of a required body movement are feasible, but we further require that every intermediate pose must be statically feasible. To this end, we generate a discritized trajectory of statically feasible poses by sampling locally from a straight-line trajectory, verifying at each point that a combination of forces exists to hold that pose statically with at least $95\%$ probability of success.
We then use SCP to maximize this probability by solving the following non-convex optimization problem:
\begin{alignat}{3}
    \label{eq:r_opt}
&\qquad \min_{\substack{s_k,\ u_k \\ \forall k \in \{1 .. N\}}} \quad && \kappa \bigg( \sum_{k=1}^N - r(s_k, u_k) \bigg) +  \\
    \label{eq:dynamics_opt}
&\ && \alpha \bigg( \sum_{k=1}^{N-1} \big(A(s_k, u_k) - s_{k+1} \big) \bigg) + \\
&\ &&  \beta \bigg(||x_{2:N} - x_{1:N-1}||_\infty + \nonumber \\
            \label{eq:normalize_opt}
&\ &&      ||y_{2:N} - y_{1:N-1}||_\infty + \\
&\ &&       ||\phi_{2:N} - \phi_{1:N-1}||_\infty \bigg) \nonumber
\end{alignat}
\begin{alignat*}{3}
&\qquad \textrm{s.t.} \quad && s_1 = s_\text{start} \\
&\qquad               \quad && s_N = s_\text{goal} \nonumber \\
&\ && \theta_\text{min} \le \theta_{k\{i\}} \le \theta_\text{max} & \quad \forall k \in \{1..N\}, i \in \{1..4\} \nonumber \\
&\ && b_\text{min} \le b_{k\{i\}} \le b_\text{max} &\quad \forall k \in \{1..N\}, i \in \{1..4\} \nonumber \\
&\ && f_\text{min} \le f_{k\{i\}} \le f_\text{max} &\quad \forall k \in \{1..N\}, i \in \{1..4\} \nonumber \\
&\ && r_\text{min} \le r(s_k,u_k) & \quad \forall k \in \{1..N\} \nonumber \\
&\ && ||s^{(w)} - s^{(w-1)}||_\infty \le \rho_s & \quad \forall w \in \{2..\text{\# iterations}\}\nonumber\\
&\ && ||u^{(w)} - u^{(w-1)}||_\infty \le \rho_u &\quad \forall w \in \{2..\text{\# iterations}\}\nonumber\\
\end{alignat*}
where the optimization variables are the state vector
\begin{equation}
    \label{eq:state_vector}
    s_k = [x_k, y_k, \phi_k, \dot{x}_k, \dot{y}_k, \dot{\phi}_k],
\end{equation}
which defines the body position $(x,y)$, orientation $\phi$, linear velocities $(\dot{x}, \dot{y})$ and angular velocity $\dot{\phi}$ at timestep $k$, and the control vector
\begin{equation}
    \label{eq:ctrl_vector}
    u_k = [f_{k\{1\}}, f_{k\{2\}}, f_{k\{3\}}, f_{k\{4\}}],
\end{equation}
which defines the control input for each boom, $f_{k\{i\}}$, equal to the prismatic force applied to boom $i$ at timestep $k$. For all variables, $(\cdot)^{(w)}$ represents the value at SCP iteration $w$.
The trajectory is executed over $N$ timesteps.

To optimize for robustness, we minimize the negative of $r(s,u)$ at all timesteps in the trajectory \eqref{eq:r_opt}.
The function $A$ in \eqref{eq:dynamics_opt} represents the non-linear dynamics of ReachBot, implemented here as a penalty constraint to aid in SCP convergence.
The normalization terms in \eqref{eq:normalize_opt} encourage smoothness. The weights $\kappa$, $\alpha$, and $\beta$ are treated as hyperparameters.

The start and goal states $s_\text{start}$ and $s_\text{goal}$ for each body movement phase are determined by the footstep planner. We apply kinematic and control constraints as they were introduced in Section~\ref{sec:feasibility}, where $b_{k\{i\}}$ is the length of boom $i$ at timestep $k$ and $\theta_{k\{i\}}$ is the angle from rotational equilibrium of boom $i$ and timestep $k$.
We additionally constrain a lower bound on robustness as defined in Section~\ref{sec:feasibility}. Finally, the trust parameters for the state $\rho_s$ and control $\rho_u$ ensure the solution to each iteration does not diverge from the local convexification.

\begin{figure}[tp]
    \centering
    \subfloat[Analysis of the body movement between Fig.~\ref{fig:level2}(a) and Fig.~\ref{fig:level2}(b).]{
        \label{fig:level1}
    \hspace{-15pt}
    \begin{subfigure}[]{0.24\textwidth}
	\includegraphics[width=\textwidth]{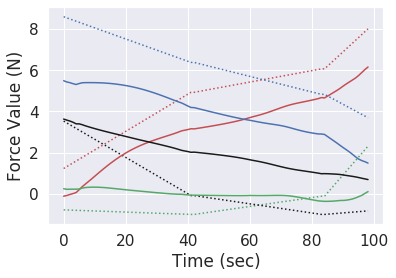}
	\label{subfig:level1_b}
    \end{subfigure}
    \begin{subfigure}[]{0.25\textwidth}
	\includegraphics[width=\textwidth]{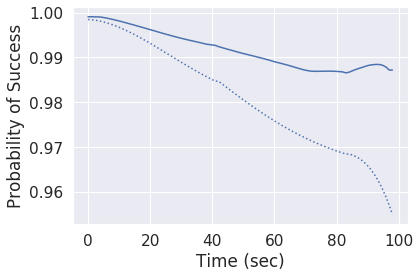}
	\label{subfig:level1_c}
	\end{subfigure} }
    \centering

	\subfloat[Analysis of the end-effector movement between Fig.~\ref{fig:level2}(b) and Fig.~\ref{fig:level2}(c).]{
		\label{fig:level15}
	\hspace{-20pt}
 	\label{subfig:level1_b}
    \begin{subfigure}[]{0.24\textwidth}
	\includegraphics[width=\textwidth]{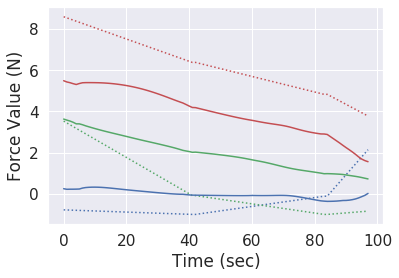}
    \end{subfigure}
    \begin{subfigure}[]{0.25\textwidth}
	\includegraphics[width=\textwidth]{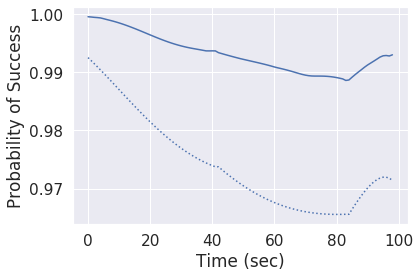}
	\end{subfigure} }
    \caption{
    (Left) applied forces (one color per boom); (right) probability of success throughout the trajectory.
    In general, the SCP-optimized trajectory (solid line) encourages less control effort than the seed trajectory (dotted line), decreasing the probability of exceeding the gripper limit surface. This corresponds to an increased probability of success in the optimized trajectory.
     }
     \label{fig:levels1and15}
     \vspace{-5pt}
\end{figure}

Fig.~\ref{fig:levels1and15}(a) shows the performance of RBP on the example body movement segment in Fig.~\ref{fig:level2}. After optimization, the controls are within actuator limits and generally smaller in magnitude, as seen in Fig.~\ref{fig:levels1and15}(a, left). A smaller pulling magnitude decreases the probability of exceeding a gripper's limit surface, as evidenced by the increased probability of success shown in Fig.~\ref{fig:levels1and15}(a, right). Both the seed and optimized trajectory are feasible with a minimum probability of success above $95\%$.

\subsection{End-effector movement phase}\label{sec:level1.5}
In the end-effector movement phase, ReachBot remains in a transitional 3-stance while it moves its free end-effector from one anchor point to another defined by the footstep planner in Section~\ref{sec:level2}. 
During end-effector movement, the cantilevered boom exerts an external wrench on the robot body, requiring additional compensation to keep ReachBot stationary. 
By keeping movement slow enough, we can ignore inertial effects and assert that the external wrench is due entirely to the cantilever. With this assertion, the external wrench is independent of the exact path taken by the end-effector, so we arbitrarily define a collision-free parabolic trajectory between the two anchor points.
We generate a seed path of feasible body poses as in Section~\ref{sec:level1}, then use SCP to maximize robustness throughout the trajectory. 

The full non-convex optimization problem is nearly identical to that in \ref{sec:level1} with only slight modifications.
First, the robustness $r(s_k, u_k)$ only considers the ellipses of the three attached points.
Second, the control vector must include torsional input at the shoulder joint in addition to prismatic forces
\begin{equation*}
    u_k = [f_{k\{i1\}}, f_{k\{i2\}}, f_{k\{i3\}}, f_{k\{j\}}, t_{k\{j\}}],
\end{equation*}
where $f_{k\{i1\}}$, $f_{k\{i2\}}$, $f_{k\{i3\}}$ are forces applied at the three anchored booms, and $f_{k\{j\}}$ and $t_{k\{j\}}$ represent the prismatic force and rotational torque, respectively, applied to moving boom $j$ at timestep $k$. %
Lastly, an external wrench term imparted by the cantilevered boom and end-effector is included in the nonlinear dynamics function $A$.

Fig.~\ref{fig:levels1and15}(b) shows a similar performance increase after optimizing end-effector movement as for body movement. We again consider an example segment from Fig.~\ref{fig:level2}. The smaller actuator output shown in Fig.~\ref{fig:levels1and15}(b, left) induces an increased probability of success shown in Fig.~\ref{fig:levels1and15}(b, right). Both the seed and optimized trajectory are verified to be feasible with a minimum probability of success above $95\%$.

\section{Results}\label{sec:results}
In this section, we introduce a naive footstep planner that does not account for limit surface constraints. We compare this naive baseline to RBP's performance in terms of success rate and computation times.
We then demonstrate body movement on a hardware platform to verify robustness throughout trajectory. 

\subsection{Simulation Results}
\begin{figure}[htp]
    \vspace{-15pt}
    \centering
    \includegraphics[width=0.5\textwidth]{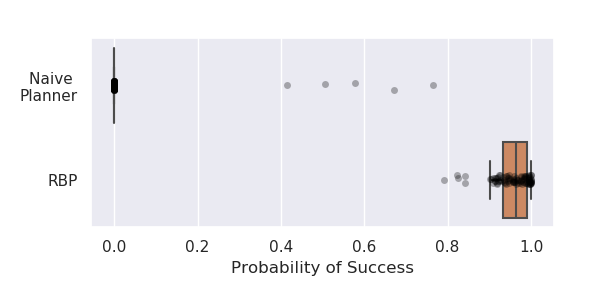}
    \caption{Probability of success of footstep planners over $100$ randomized trials. The naive baseline planner does not account for limit surfaces, leading to a low success rate due to gripper failure, though it does succeed in some instantiations. RBP returns footstep paths with over $90\%$ probability of success in $86$ trials out of $100$ and paths with over $95\%$ success rate in more than half of the trials.}
    \label{fig:baseline}
\end{figure}

To our knowledge, there are no existing motion planners that account for the stochastic limit surfaces characteristic of microspine grippers. We construct a naive footstep planner that imitates the graph traversal algorithm presented in Section~\ref{sec:level2} but without considering the feasibility of the poses due to gripper limitations. 
Fig.~\ref{fig:baseline} shows the results of $100$ paths generated by both footstep planners for environments with randomized anchor point placement and limit surface magnitude and orientation. 
To minimize distance-to-goal, the naive planner greedily selects the farthest anchor point in reach, resulting in a plan with fewer regrasps. In general, fewer regrasps corresponds to fewer chances for failure. However, the overwhelming superiority of RBP highlights the importance of considering limit surfaces: a path with more regrasps that ensures higher robustness per grasp performs better overall than simply minimizing number of regrasps.

Table~\ref{table} provides the average compute time of RBP's footstep planner over $100$ randomized iterations as compared with the naive planner. As the naive planner produces so few feasible footstep paths, we cannot compare computation time of the continuous portions of the trajectory to a baseline. The computation time for the initial seed trajectory as well as additional time to perform SCP optimization are listed in Table~\ref{table} for individual body and end-effector movement segments. 
The total computation time for the multistep trajectory plan is, for our unoptimized implementation, on the order of minutes owing to the highly nonlinear nature of the optimization and our corresponding conservative trust region definition. We note, however, that the computation time is compatible (even accounting for replanning) with the intended quasistatic deployment of ReachBot where this duration is comparable to the implementation of a single body movement segment.

\begin{table}[h!]
\centering
\begin{tabular}{||c || c | c||} 
 \hline
  & RBP planning & Naive planning \\ 
  & time (s) & time (s)\\ [0.5ex] 
 \hline\hline
 footstep plan & 66.19 & 0.0123 \\ 
 \hline
 seed for body mvmt & 0.0839 & N/A  \\
 \hline
 SCP for body mvmt & 10.80 &  N/A  \\
 \hline
 seed for end-eff. mvmt & 0.0417 & N/A   \\
 \hline
 SCP for end-eff. mvmt & 50.09 & N/A  \\ 
 \hline
\end{tabular}
\caption{Computational time for each segment of the motion plan.}
\label{table}
\end{table}

\subsection{Experimental Results}
In order to validate the trajectory produced by RBP, we built a ReachBot prototype with a small robot body and elastic cords as its long, extendable booms. There are four motors driving four independent winches controlling both the lengths and tensions of the booms. The anchors are equipped with a single axis load cell mounted on a hinge which allows them to freely rotate. Since the booms are rigidly attached to the anchors, this setup can only be used to test body movement trajectories. The full experimental setup is shown in Fig.~\ref{fig:hardware}, with a close-up view of an anchor in Fig.~\ref{fig:limit_surface}.

\begin{figure}[hbp]
\vspace{-15pt}
    \centering
    \includegraphics[width=0.4\textwidth]{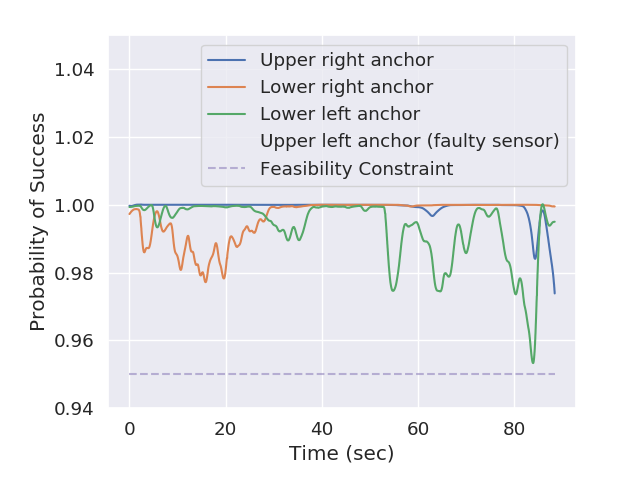}
    \caption{Probability of grasp maintenance for each anchor on our hardware platform while implementing a single body movement trajectory.
    }
    \label{fig:experiment}
    \vspace{-10pt}
\end{figure}

To validate simulation results, we execute a body movement trajectory on our hardware prototype. Using the force and angle at each anchor, we calculate the probability of success of each representative grasp based on the predefined limit surface parameters. The results are shown in Figure~\ref{fig:experiment} with a timestep of $1$s. Notable trends include the increasing probability of success at the lower right anchor as the boom angle approaches the ellipse normal direction ($\psi = 0$). Similarly, we see the probability of success for the lower left anchor decrease as the boom angle diverts from the ellipse's normal.

\section{Conclusion}
In this paper, we presented ReachBot Planner (RBP), a motion planner for a climbing robot that incorporates stochastic limit surfaces at every anchor point. We developed a convex metric for robustness based on an ellipsoidal model of a microspine limit surface and applied it to a decomposed motion planning paradigm. First, RBP determines a sequence of stances wherein there exists a feasible transition between each pair of adjacent stances. Then, it uses sampling-based motion planning to generate a feasible seed trajectory for each body and end-effector movement segment. For each segment, RBP uses sequential convex programming to maximize probability of success throughout the trajectory. We verified RBP both in simulation and on a hardware platform and compared it to a naive baseline to prove the importance of considering grasp limit surfaces in high-level planning.

Our eventual goal is to design, build, and field-test a 3D ReachBot prototype. Extending RBP to 3D should be straightforward as the convex formulation of the robustness metric is preserved in 3D. However, full knowledge of available grasp points is unrealistic in a real-world setting, so we are concurrently developing perception strategies to quantify grasp sites in an unfamiliar environment. 
Additionally, we will develop a metric for robustness to dynamic failure and perform experiments on the hardware platform presented in this paper before deploying it on a 3D robot.


\section*{Acknowledgements}
Support for this work was provided by NASA under the NIAC program and by the Air Force under an STTR award with Altius Space Machines. S. Newdick is supported by NASA NSTGRO and T. G. Chen is supported by NSF GRFP.


\bibliographystyle{IEEEtran}
\bibliography{ASL_papers,main}

\newcommand{\noopsort}[1]{} \newcommand{\printfirst}[2]{#1}
  \newcommand{\singleletter}[1]{#1} \newcommand{\switchargs}[2]{#2#1}
\begin{thebibliography}{10}
\providecommand{\url}[1]{#1}
\csname url@rmstyle\endcsname
\providecommand{\newblock}{\relax}
\providecommand{\bibinfo}[2]{#2}
\providecommand\BIBentrySTDinterwordspacing{\spaceskip=0pt\relax}
\providecommand\BIBentryALTinterwordstretchfactor{4}
\providecommand\BIBentryALTinterwordspacing{\spaceskip=\fontdimen2\font plus
\BIBentryALTinterwordstretchfactor\fontdimen3\font minus
  \fontdimen4\font\relax}
\providecommand\BIBforeignlanguage[2]{{%
\expandafter\ifx\csname l@#1\endcsname\relax
\typeout{** WARNING: IEEEtran.bst: No hyphenation pattern has been}%
\typeout{** loaded for the language `#1'. Using the pattern for}%
\typeout{** the default language instead.}%
\else
\language=\csname l@#1\endcsname
\fi
#2}}

\bibitem{NRC2011}
{National Research Council}, ``Decadal survey vision and voyages for planetary
  science in the decade 2013--2022,'' {National Academy Press}, Tech. Rep.,
  2011.

\bibitem{LapotreORourkeEtAl2020}
M.~G.~A. Lap\^{o}tre, J.~G. O’Rourke, L.~K. Schaefer, K.~L. Siebach,
  C.~Spalding, S.~M. Tikoo, and R.~D. Wordsworth, ``Probing space to understand
  earth,'' \emph{Nature Reviews Earth \& Environment}, vol.~1, pp. 170--181,
  2020.

\bibitem{SchneiderBylardEtAl2022}
S.~Schneider, A.~Bylard, T.~G. Chen, P.~Wang, M.~R. Cutkosky, and M.~Pavone,
  ``{ReachBot:} {A} small robot for large mobile manipulation tasks,'' in
  \emph{{IEEE Aerospace Conference}}, 2022, in press.

\bibitem{ChenMillerEtAl2022}
T.~G. Chen, B.~Miller, C.~Winston, S.~Schneider, A.~Bylard, M.~Pavone, and
  M.~R. Cutkosky, ``{ReachBot:} {A} small robot with exceptional reach for
  rough terrain,'' in \emph{{Proc.\ IEEE Conf.\ on Robotics and Automation}},
  2022, press.

\bibitem{LeclercWilsonEtAl2017}
C.~Leclerc, L.~Wilson, M.~A. Bessa, and S.~Pellegrino, ``Characterization of
  ultra-thin composite triangular rollable and collapsible booms,'' in
  \emph{{AIAA SPACE Conferences \& Exposition}}, 2017.

\bibitem{HebertBajracharyaEtAl2015}
P.~Hebert, M.~Bajracharya, J.~Ma, N.~Hudson, A.~Aydemir, J.~Reid, C.~Bergh,
  J.~Borders, M.~Frost, M.~Hagman, J.~Leichty, P.~Backes, B.~Kennedy,
  P.~Karplus, B.~Satzinger, K.~Byl, K.~Shankar, and J.~Burdick, ``Mobile
  manipulation and mobility as manipulation—{Design} and algorithms of
  {RoboSimian},'' \emph{{Journal of Field Robotics}}, vol.~32, no.~2, pp.
  255--274, 2015.

\bibitem{Wilcox2012}
B.~H. Wilcox, ``{ATHLETE}: {A} limbed vehicle for solar system exploration,''
  in \emph{{IEEE Aerospace Conference}}, 2012.

\bibitem{HeverlyMatthewsEtAl2010}
M.~Heverly, J.~Matthews, M.~Frost, and C.~{McQuin}, ``Development of the
  {Tri-ATHLETE} lunar vehicle prototype,'' in \emph{Aerospace Mechanisms
  Symposium}, 2010.

\bibitem{Parness2017}
A.~Parness, N.~Abcouwer, C.~Fuller, N.~Wiltsie, J.~Nash, and B.~Kennedy,
  ``{LEMUR 3:} {A} limbed climbing robot for extreme terrain mobility in
  space,'' in \emph{{Proc.\ IEEE Conf.\ on Robotics and Automation}}, 2017.

\bibitem{Aceituno-CabezasMastalliEtAl2017}
B.~Aceituno-Cabezas, C.~Mastalli, H.~Dai, M.~Focchi, A.~Radulescu, D.~G.
  Caldwell, J.~Cappelletto, J.~C. Grieco, G.~Fernandez-Lopez, and C.~Semini,
  ``Simultaneous contact, gait, and motion planning for robust multilegged
  locomotion via mixed-integer convex optimization,'' \emph{{IEEE Robotics and
  Automation Letters}}, vol.~3, no.~3, pp. 2531--2538, 2017.

\bibitem{AhnChaeEtAl2018}
M.~S. Ahn, H.~Chae, and D.~W. Hong, ``Stable, autonomous, unknown terrain
  locomotion for quadrupeds based on visual feedback and mixed-integer convex
  optimization,'' in \emph{{IEEE/RSJ Int.\ Conf.\ on Intelligent Robots \&
  Systems}}, 2018.

\bibitem{WinklerBellicosoEtAl2018}
A.~W. Winkler, C.~D. Bellicoso, M.~Hutter, and J.~Buchli, ``Gait and trajectory
  optimization for legged systems through phase-based end-effector
  parameterization,'' \emph{{IEEE Robotics and Automation Letters}}, vol.~3,
  pp. 1560--1567, 2018.

\bibitem{LinZhangEtAl2019}
X.~Lin, J.~Zhang, J.~Shen, G.~Fernandez, and D.~W. Hong, ``Optimization based
  motion planning for multi-limbed vertical climbing robots,'' in
  \emph{{IEEE/RSJ Int.\ Conf.\ on Intelligent Robots \& Systems}}, 2019.

\bibitem{PosaCantuEtAl2014}
M.~Posa, C.~Cantu, and R.~Tedrake, ``A direct method for trajectory
  optimization of rigid bodies through contact,'' \emph{{Int.\ Journal of
  Robotics Research}}, vol.~33, no.~1, pp. 69--81, 2014.

\bibitem{LaurenziHoffmanEtAl2018}
A.~Laurenzi, E.~M. Hoffman, and N.~G. Tsagarakis, ``Quadrupedal walking motion
  and footstep placement through linear model predictive control,'' in
  \emph{{IEEE/RSJ Int.\ Conf.\ on Intelligent Robots \& Systems}}, 2018.

\bibitem{SintovAvramovichEtAl2011}
A.~Sintov, T.~Avramovich, and A.~Shapiro, ``Design and motion planning of an
  autonomous climbing robot with claws,'' \emph{{Robotics and Autonomous
  Systems}}, vol.~59, no.~11, pp. 1008--1019, 2011.

\bibitem{HauserBretlEtAl2005}
K.~Hauser, T.~Bretl, and J.-C. Latombe, ``Non-gaited humanoid locomotion
  planning,'' in \emph{{IEEE RAS Int.\ Conf.\ on Humanoid Robots}}, 2005.

\bibitem{SchulmanDuanEtAl2014}
J.~Schulman, Y.~Duan, J.~Ho, A.~Lee, I.~Awwal, H.~Bradlow, J.~Pan, S.~Patil,
  K.~Goldberg, and P.~Abbeel, ``Motion planning with sequential convex
  optimization and convex collision checking,'' \emph{{Int.\ Journal of
  Robotics Research}}, vol.~33, no.~9, pp. 1251--1270, 2014.

\bibitem{ZhangLatomb2013}
R.~Zhang and J.-C. Latombe, ``Capuchin: A free-climbing robot,'' \emph{{Int.\
  Journal of Advanced Robotic Systems}}, vol.~10, no.~4, 2013.

\bibitem{AsbeckKimEtAl2006}
A.~T. Asbeck, S.~Kim, M.~R. Cutkosky, W.~R. Provancher, and M.~Lanzetta,
  ``Scaling hard vertical surfaces with compliant microspine arrays,''
  \emph{{Int.\ Journal of Robotics Research}}, vol.~25, no.~12, pp. 1165--1179,
  2006.

\bibitem{RuotoloRoigEtAl2019}
W.~Ruotolo, F.~S. Roig, and M.~R. Cutkosky, ``Load-sharing in soft and spiny
  paws for a large climbing robot,'' \emph{{IEEE Robotics and Automation
  Letters}}, vol.~4, no.~2, pp. 1439--1446, 2019.

\bibitem{JiangWangEtAl2018}
H.~Jiang, S.~Wang, and M.~R. Cutkosky, ``Stochastic models of compliant spine
  arrays for rough surface grasping,'' \emph{{Int.\ Journal of Robotics
  Research}}, vol.~37, no.~7, pp. 669--687, 2018.

\bibitem{WangJiangEtAl2019}
S.~Wang, H.~Jiang, T.~Myung~Huh, D.~Sun, W.~Ruotolo, M.~Miller, W.~R.~T.
  Roderick, H.~S. Stuart, and M.~R. Cutkosky, ``Spinyhand: {Contact} load
  sharing for a human-scale climbing robot,'' \emph{Journal of Mechanisms and
  Robotics}, vol.~11, no.~3, 2019.

\bibitem{Jiang2017}
H.~Jiang, ``Surface grasping with bio-inspired, opposed adhesion,'' Ph.D.
  dissertation, {Stanford University}, 2017.

\end{thebibliography}


\end{document}